\documentclass[conference]{IEEEtran}
\usepackage[T1]{fontenc}
\IEEEoverridecommandlockouts
\usepackage{cite}
\usepackage{amsmath,amssymb,amsfonts}
\usepackage{algorithmic}
\usepackage{graphicx}
\usepackage{textcomp}
\usepackage{xcolor}
\usepackage{hyperref}
\usepackage{multirow}

\usepackage{graphicx}
\usepackage{subcaption}
\usepackage{caption}

\def\BibTeX{{\rm B\kern-.05em{\sc i\kern-.025em b}\kern-.08em
    T\kern-.1667em\lower.7ex\hbox{E}\kern-.125emX}}
\begin{document}

\title{Hardware-Aware Feature Extraction Quantisation for Real-Time Visual Odometry on FPGA Platforms \\
\thanks{The work presented in this paper was supported by the programme ``Excellence initiative -- research university'' and 16.16.120.773 for the AGH University of Krakow.}
}

\author{\IEEEauthorblockN{Mateusz Wasala, Mateusz Smolarczyk, Michał Danilowicz, Tomasz Kryjak \textit{Senior Member IEEE}}

\IEEEauthorblockA{
\textit{Embedded Vision Systems Group, Department of Automatic Control and Robotics,} \\
\textit{AGH University of Krakow, Poland}}
E-mail: \textit{mateusz.wasala@agh.edu.pl},
\textit{mtsmolar@student.agh.edu.pl}, 
\textit{danilowi@agh.edu.pl},
\textit{tomasz.kryjak@agh.edu.pl}
} 

\maketitle

\begin{abstract}
Accurate position estimation is essential for modern navigation systems deployed in autonomous platforms, including ground vehicles, marine vessels, and aerial drones. In this context, Visual Simultaneous Localisation and Mapping (VSLAM) -- which includes Visual Odometry -- relies heavily on the reliable extraction of salient feature points from the visual input data. 
In this work, we propose an embedded implementation of an unsupervised architecture capable of detecting and describing feature points. 
It is based on a quantised SuperPoint convolutional neural network.
Our objective is to minimise the computational demands of the model while preserving high detection quality, thus facilitating efficient deployment on platforms with limited resources, such as mobile or embedded systems. 
We implemented the solution on an FPGA System-on-Chip (SoC) platform, specifically the AMD/Xilinx Zynq UltraScale+, where we evaluated the performance of Deep Learning Processing Units (DPUs) and we also used the Brevitas library and the FINN framework to perform model quantisation and hardware-aware optimisation. This allowed us to process $640 \times 480$ pixel images at up to 54 fps on an FPGA platform, outperforming state-of-the-art solutions in the field.
We conducted experiments on the TUM dataset to demonstrate and discuss the impact of different quantisation techniques on the accuracy and performance of the model in a visual odometry task.
\end{abstract}


\section{Introduction}


Position estimation is a key issue in many modern applications, such as navigation of autonomous robots, including land vehicles, marine vessels or drones, augmented reality (AR) and advanced driver assistance systems (ADAS). 
Different methods are used to determine the position and movement of a robot, depending on the environmental conditions. 
The most commonly used technology is the Global Navigation Satellite System (GNSS), but its performance can be limited in confined, urban environments or in the presence of external interference. 
To increase the reliability and robustness of localisation algorithms, data from sensors directly mounted on the robot, such as cameras, lidars, radars or inertial measurement units (IMUs), are increasingly being used. 
Among the different types of sensors, cameras are inexpensive and can provide rich information about the environment, ensuring reliable and accurate position estimation. 


Simultaneous localisation and mapping (SLAM) is a fundamental task for many applications for moving robots, such as terrain exploration and indoor navigation.
Visual odometry (VO) is a method that enables the estimation of a robot's trajectory or position by calculating the displacement between two successive images of the environment in which the robot is operating.
Consequently, it is a key component of SLAM systems.
The most popular VO solution relies on feature point detection and matching.
In this approach, the set of operations typically includes: feature extraction, feature matching, outlier rejection, motion estimation, and local optimisation based on bundle adjustment (BA). 
For VO systems that require high real-time performance, it is important to efficiently extract robust features and descriptors for matching.
Traditional feature extraction methods, such as SIFT \cite{sift2004}, SURF \cite{bay2006surf} and ORB \cite{orb}, remain widely used and have formed the foundation of classic visual odometry systems for many years. 
However, in recent years, approaches based on deep neural networks have become increasingly popular and are now being used in modern localisation and mapping systems due to their higher accuracy, robustness to noise, and overall improved performance. Despite these advantages, the embedded implementation of deep learning-based methods remains a significant challenge, primarily due to limited computational resources, power constraints, and the need for real-time performance on edge devices.

In this paper, we present an analysis of the feasibility of implementing a network for feature point detection and descriptor on an embedded system.
We examine the use of various quantisation methods in the feature extraction process, taking into account hardware limitations and available acceleration mechanisms.
Experiments were conducted using the SuperPoint network \cite{detone2018superpoint} as a case study, in the context of application to the task of visual odometry.
We present two approaches to the hardware implementation of this model. One using the Vitis AI tool and the other based on the Brevitas and FINN frameworks.

The main contributions of this work are summarised as follows:
\begin{itemize}
    \item Presentation and analysis of various methods of quantisation of neural networks, with particular attention to their impact on the quality of feature extraction in the task of visual odometry.
    
    \item Implementation and optimisation of a quantised version of the SuperPoint network, designed for efficient feature point detection and description, with an emphasis on maintaining high detection quality with limited computational resources.

    
    \item The proposed design has been configured for two parallelisation variants and evaluated on Kria KV260 and ZCU102 FPGA boards. It achieves real-time performance of 27 FPS and 54 FPS respectively, outperforming state-of-the-art solutions in the field.

    \item The experimental evaluation of model performance and post-training quantisation quality was carried out to demonstrate the practical benefits of the proposed methods employed in the context of real-world visual odometry applications.
\end{itemize}

The remainder of the article is organised as follows. Section \ref{sec:previous_works} reviews existing solutions. Section \ref{sec:proposed_method}, we describe the SuperPoint network architecture, the quantisation tools used, and the visual odometry pipeline. Section \ref{sec:hardware} details the hardware implementation, and Section \ref{sec:evaluation} presents the evaluation results. Finally, the paper concludes with a summary and discussion of future research directions.

\section{Previous works}
\label{sec:previous_works}


A fundamental component of most VSLAM and VO systems is the process of detection and feature matching. 
On the one hand, it determines the overall effectiveness of the system; on the other hand, it represents the most computationally demanding stage.
As a result, accelerating this stage is particularly important in the context of implementation on embedded platforms, where computing resources are limited.

The approaches described in the literature can be divided into two categories: classical methods and those based on artificial intelligence.
The first group includes algorithms such as SIFT \cite{sift2004}, SURF \cite{bay2006surf}, and ORB \cite{orb}, which are still widely used in visual odometry. 
Although effective, classical methods are often sensitive to changes in illumination, texture, and the presence of noise, which limits their applicability in more complex or dynamic environments.

In recent years, deep learning-based methods, which have proven more effective in addressing these challenges, have experienced a marked increase in popularity. 
Position estimation using neural networks can be implemented in two distinct ways. 
The first is a detector-free approach,in which the neural network is trained to directly map the input data (e.g., image sequences) to the estimated position and orientation of the camera. 
The second is a hybrid approach, where deep models are used to replace specific components of the classical visual odometry pipeline, such as feature point detection, feature point matching, and geometric transformations estimation.

A variety of architectural approaches have been employed to address the aforementioned tasks. Convolutional Neural Networks (CNNs) \cite{detone2018superpoint}, Graph Neural Networks (GNNs) \cite{superglue2020, chen2021learning}, and Transformers \cite{sun2021loftr} represent three examples of deep learning models. 
These approaches have been shown to improve accuracy and robustness, as well as to enhance adaptability to various environmental conditions.
Despite the high quality of their results, neural network-based solutions are characterised by high computational complexity, which significantly limits their applicability in real-time systems operating on resource-constrained platforms, such as mobile robots and handheld devices.

In this work, we focus on feature point detection, using the SuperPoint convolutional neural network \cite{detone2018superpoint} as a representative example.
SuperPoint is one of the most widely adopted approaches in visual odometry.
It is a convolutional, self-supervising neural network that simultaneously generates feature points and their corresponding descriptors. 
This network is often combined with the SuperGlue \cite{superglue2020} network to match feature points.

Several recent studies have addressed the acceleration and quantisation of the SuperPoint convolutional neural network.
In work \cite{xu2020}, the Deep Learning Processor Unit (DPU) was applied and the CNN backbone of SuperPoint was quantised to 8-bit fixed-point representation, resulting in a negligible accuracy loss. 
Both the CNN backbone and the post-processing operations were quantised to 8-bit fixed-point numbers. 
Additionally, the base of the exponential function in the Softmax operation was changed from $e$ to $2$.
CNN inference is performed using the Xilinx DPU accelerator, which is a hardware IP (provided by Xilinx FPGA for accelerating neural networks) implemented in the programmable logic (PL) of the ZCU102 FPGA chip. 
The Softmax and normalisation stages are also implemented on the hardware accelerators proposed by the authors, which are integrated within the same PL part. 
In contrast, the Non-Maximum Suppression (NMS) operations and feature point matching are implemented on the CPU, i.e., within the processing system (PS) section of the chip. 
The entire feature point extraction method operates in real time, achieving a performance of 20 frames per second on the ZCU102 embedded FPGA platform at a resolution of $640 \times 480$ pixels.

The paper \cite{liu2022} proposes a real-time hardware accelerator for feature point extraction based on the SuperPoint network architecture, implemented on an FPGA chip and referred to as MobileSP.
MobileSP is a simplified variant of the original SuperPoint architecture. 
It features a reduced number of convolutional layers in the backbone and omits max-pooling operations, thereby decreasing computational complexity and the total number of network parameters.
The neural network is implemented using an 8-bit DPU.  
The other parts, including Softmax, NMS and the Descriptor Decoder, are implemented as custom hardware modules in the PL part, and the processing system (PS) is used to coordinate the operation and data exchange between the different modules.
The design was deployed and evaluated on a Xilinx ZCU104 FPGA board.
During testing, images were transmitted to the FPGA via Ethernet, and the detected feature points were retrieved for use in an ORB-SLAM2-based visual odometry system, where SuperPoint was substituted as the feature extractor.
The accelerator achieved real-time performance of 42 frames per second, with a low absolute trajectory error (ATE) of 1.82 cm on the TUM dataset.

In the work \cite{Yin2024}, the authors proposed a feature extraction accelerator based on a convolutional neural network, characterised by low hardware overhead and end-to-end processing capability. 
In order to achieve full acceleration of the lightweight version of the SuperPoint network, a dedicated, fully pipelined hardware architecture was designed. 
The default network architecture was modified by reducing the number of filters in the convolution layers: by a factor of 1/8 in the initial layers, 1/4 in the intermediate layers and 1/2 in the final layers, relative to the original architecture.
These modifications resulted in a significant reduction in computational complexity.
The proposed accelerator, implemented on the Xilinx ZCU102 FPGA platform, enables real-time processing of $640 \times 480$ resolution images at 20 frames per second and a clock frequency of 200 MHz, while achieving a high energy efficiency of 58.2 mJ per frame.

Existing work in the literature primarily focusses on the use of 8-bit DPUs to accelerate neural network computations, low-level implementation of selected network layers in FPGA programmable logic, or architectural modifications aimed at enabling real-time processing.
In this paper, we propose an alternative approach -- instead of modifying the architecture itself, we focus on optimising the computations through various quantisation strategies. 
This enables a significant reduction in computational and energy requirements, while maintaining high model accuracy.

\section{The proposed method}
\label{sec:proposed_method}

\begin{figure*}[!ht]
  \centering
  \begin{subfigure}[t]{0.67\textwidth}
    \centering
    \includegraphics[width=\linewidth]{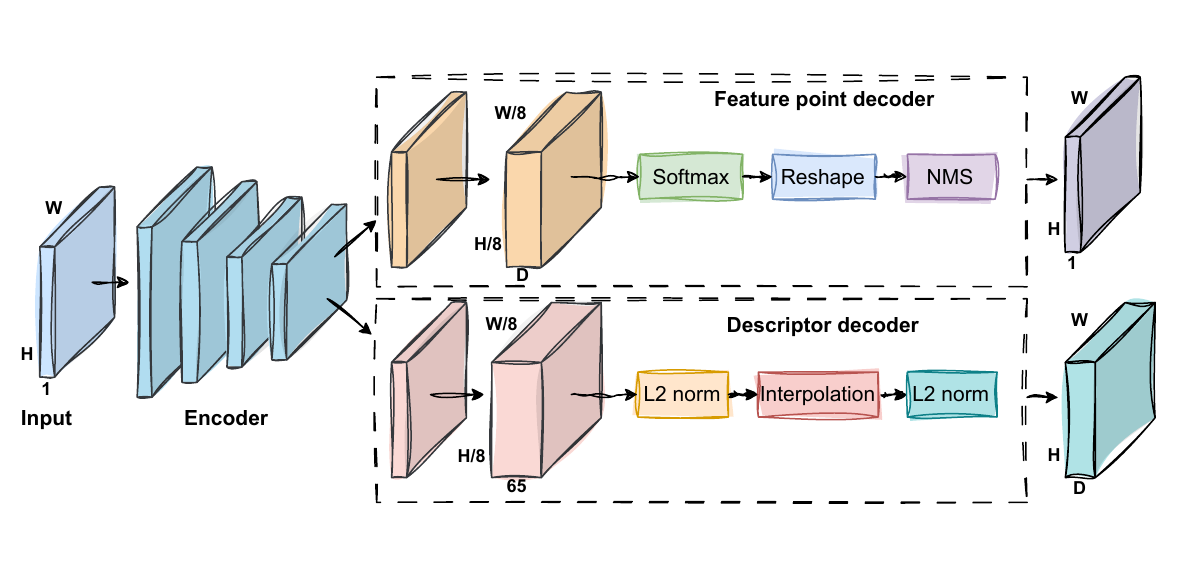}
    \caption{SuperPoint network architecture.}
    \label{fig:superpoint_arch}
  \end{subfigure}
  \begin{subfigure}[t]{0.2\textwidth}
    \centering
    \includegraphics[width=\linewidth]{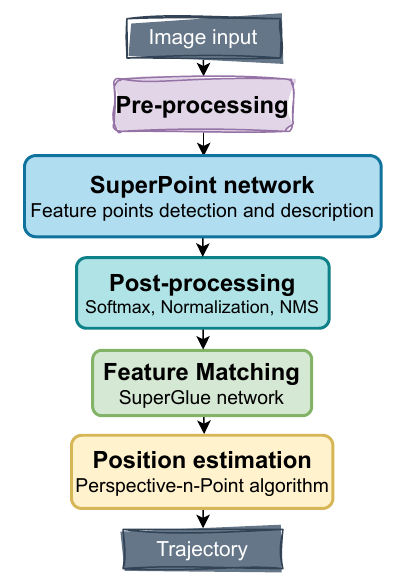}
    \caption{Visual odometry pipeline.}
    \label{fig:odometry_pipeline}
  \end{subfigure}

  \caption{Schemes representing the main elements of the proposed method: \subref{fig:superpoint_arch} Convolution neural network architecture called SuperPoint \cite{detone2018superpoint}, \subref{fig:odometry_pipeline} Visual odometry pipeline using the SuperPoint detector.}
  \label{fig:main}
\end{figure*}

This paper analyses different quantisation methods of the SuperPoint neural network \cite{detone2018superpoint}, used to detect and describe feature points in an image in the context of visual odometry. 

\subsection{SuperPoint}

SuperPoint is a convolutional neural network for detecting feature points and generating their descriptors. It takes as input a greyscale image with dimensions (1, W, H). Figure \ref{fig:superpoint_arch} visualises the structure of the SuperPoint network.
It consists of three main components:
\begin{itemize}
    \item encoder, responsible for extracting relevant image features,
    \item descriptor decoder, generating descriptions for each pixel,
    \item feature point decoder, which creates a map of feature point locations.
\end{itemize}


\subsubsection{Encoder} 
The main part of the SuperPoint network is a shared encoder that is responsible for extracting relevant features from the input image. These features are then transmitted to two distinct decoders: one responsible for feature point detection and the other for generating their descriptors. The encoder is composed of four convolution blocks, each of which contains two convolution layers with $3\times3$ filters, followed by a MaxPooling operation with a $2\times2$ window. 
The output of the encoder is a $(128, W/8, H/8)$ tensor, which contains a dense representation of the image features and forms the basis for further processing by the two decoders.

\subsubsection{Feature point decoder}

The feature point detection process begins with two convolution layers that generate a $(65, W/8, H/8)$ tensor. 
This tensor encodes the probability of detecting a feature point in each of the 64 positions within a $8\times8$ grid, while the 65th channel corresponds to the absence of a feature point (i.e., the background class).
A Softmax function is then applied across the channel dimension, converting the raw responses into a probability distribution.
Subsequently, the 65th channel is discarded, and the remaining 64 channels are used for upsampling. This operation restores the spatial resolution to that of the original input image, resulting in an output tensor of shape $(1, W, H)$.
In the final step, the Non-Maximum Suppression (NMS) algorithm is used to eliminate points with lower response in the local neighbourhood, leaving only local maxima. 
Additionally, feature points located too close to the image boundaries are discarded to avoid unstable detections.

\subsubsection{Descriptor decoder}

As with the previously described decoder, the processing begins with two convolution layers that generate a tensor of dimension $(256,W/8,H/8)$. 
This is followed by an initial feature normalisation step, which stabilises the magnitude of the vectors in the feature space.
Next, bilinear interpolation is applied to upscale the tensor to the resolution of the input image, resulting in a dense descriptor map. 
Finally, a second normalisation step ensures consistent descriptor scaling, making the output robust to local variations in contrast and illumination.
The final output of the decoder is a tensor of shape $(256, W, H)$, containing dense descriptors that represent local image features at each pixel location.

\subsection{Visual Odometry}

The SuperPoint neural network, as discussed in the paper, was used for the task of visual odometry, which comprises the following components:

\begin{itemize}
    \item Feature point detection and descriptor using the SuperPoint network.
    \item Matching of feature points using the SuperGlue network \cite{superglue2020}.
    \item Position estimation from greyscale and depth images.
\end{itemize}

Figure \ref{fig:odometry_pipeline} shows the scheme of the visual odometry pipeline using the SuperPoint detector.
Before feature point detection, distortion removal was performed on images from the TUM RGB-D SLAM Dataset \cite{tum_dataset}.
The SuperPoint network is then used for detection, taking into account the appropriate precision of the model weights.
The subsequent step involves utilising the SuperGlue graph neural network to match feature points. The default parameters are utilised as a starting point.
We used the OpenCV implementation of Perspective-n-Point (solvePnP) with all the matches to compute the transformation matrix. 
We used greyscale and depth images for position estimation. 
The depth image allows us to estimate the correct scale.
We did not use additional local optimisation algorithms such as Bundle Adjustment to correctly evaluate the effect of quantisation on feature point detection and position estimation.

\subsection{Quantisation-aware training}

Quantisation-aware training (QAT) was performed to prepare the model for hardware implementation.
This is one way to reduce the memory and computational complexity of the model in which the model parameters and activations are represented in integer or fixed-point format as opposed to the typical floating-point representation.
There are two primary reasons for this procedure.
First, a smaller number of bits per representation can be used: 8, 4, 2, or even 1, which significantly affects the hardware computation cost.
Second, addition and multiplication operations in an integer (or equivalently fixed-point) representation are several times less costly (in terms of digital chip area and power consumption) than in a floating-point representation \cite{barrois2017energy}.

For QAT, the Brevitas library \cite{brevitas} was used, implementing uniform affine quantisation \cite{jacob2018quantization}.
After each convolution layer, activation requantisation was introduced to avoid uncontrolled growth of the number of bits per representation.
The quantisation scales were kept at floating-point precision and were common per activation tensor, while for the model weights, each channel had an individually selected scale.
The zero-point quantisation parameter was set to $0$ due to a simpler hardware implementation.

\begin{table*}[!ht]
\centering
\caption{A comparison of repeatability detection, localisation error, and homography estimation metrics for images with a resolution of $640\times 480$ px on a set of HPatches has been conducted.
For all metrics, the highest possible accuracy values are desired, with the exception of localization error, for which the lowest possible values are preferred.}
\label{tab:detector_metrics}
\begin{tabular}{lccc|ccccc|}
\cline{5-9}
\multicolumn{1}{l}{} &
  \multicolumn{1}{l}{} &
  \multicolumn{1}{l}{} &
  \multicolumn{1}{l|}{} &
  \multicolumn{5}{c|}{\textbf{Precision}} \\ \hline
\multicolumn{1}{|c|}{\textbf{Detector Metrics}} &
  \multicolumn{1}{l|}{\textbf{FCCM’20 \cite{xu2020}}} &
  \multicolumn{1}{l|}{\textbf{TCASI’22 \cite{liu2022}}} &
  \multicolumn{1}{l|}{\textbf{APCCAS'24 \cite{Yin2024}}} &
  \multicolumn{1}{c|}{\textbf{\begin{tabular}[c]{@{}c@{}}Baseline \\ (FP32 \cite{detone2018superpoint}) \end{tabular}}} &
  \multicolumn{1}{c|}{\textbf{INT8}} &
  \multicolumn{1}{c|}{\textbf{INT4}} &
  \multicolumn{1}{c|}{\textbf{INT3}} &
  \textbf{\begin{tabular}[c]{@{}c@{}}Mixed precision \\ (INT 4-2-4) \end{tabular}} \\ \hline
\multicolumn{1}{|l|}{Repeatability} &
  \multicolumn{1}{c|}{0.52**} &
  \multicolumn{1}{c|}{0.53} &
                      0.53 &
  \multicolumn{1}{c|}{0.574} &
  \multicolumn{1}{c|}{0.565} &
  \multicolumn{1}{c|}{0.547} &
  \multicolumn{1}{c|}{0.522} &
                        0.50 \\
\multicolumn{1}{|l|}{Localization Error} &
  \multicolumn{1}{c|}{-} &
  \multicolumn{1}{c|}{1.13} &
                    1.28 &
  \multicolumn{1}{c|}{1.17} &
  \multicolumn{1}{c|}{1.18} &
  \multicolumn{1}{c|}{1.24} &
  \multicolumn{1}{c|}{1.34} &
  1.39 \\
\multicolumn{1}{|l|}{Homography Estimation (e=3)} &
  \multicolumn{1}{c|}{0.56**} &
  \multicolumn{1}{c|}{0.83} &
  0.68 &
  \multicolumn{1}{c|}{0.82} &
  \multicolumn{1}{c|}{0.81} &
  \multicolumn{1}{c|}{0.76} &
  \multicolumn{1}{c|}{0.75} &
  0.71 \\ \hline
\end{tabular}

**: average value calculated based on original data 
\end{table*}

\section{Embedded implementation}
\label{sec:hardware}


The application of neural networks in feature extraction leads to an increase in computational complexity, which can result in slower algorithmic performance. 
To ensure real-time operation of deep learning-based visual odometry algorithms (at least 20-25 FPS), it is necessary to accelerate them using dedicated hardware. 
Real-time capability is crucial for vision systems to respond effectively to dynamically changing environmental conditions.
For this study, two FPGA platforms were selected: the Xilinx Kria KV260 Vision AI Starter Kit and the Xilinx ZCU102 development board. 
Both platforms are equipped with the Zynq UltraScale+ MPSoC chip, which integrates a classical processor (PS -- Processing System) based on a quad-core ARM Cortex-A53 and a programmable logic part (PL -- Programmable Logic), where a Deep Learning Processing Unit (DPU) accelerator can be implemented. 
The Kria KV260 platform is optimised for rapid deployment in vision-based AI applications, while the ZCU102 provides a more flexible development environment for prototyping and performance evaluation.
Two solutions were tested to implement the deep neural network, using the Vitis AI tool, which offers the use of a dedicated DPU, and the FINN framework to implement the model in the PL part of the MPSoC.

\subsection{Quantisation and Model Training Process}

We analysed different variants of neural network quantisation, including quantisation to 8, 4 and 3 bits, as well as a mixed precision approach (4-2-4), allowing a compromise between quality and computational efficiency.
For the mixed precision (4-2-4) variant, we used 4-bit precision in the first convolutional layer of the encoder, as well as in the last ReLU and convolutional layers in the Feature Point Decoder and Descriptor Decoder, respectively. For the other layers of the network, we used 2-bit precision. 
For training, we used a pretrained floating-point model provided by the authors of the SuperPoint network.
Quantisation-aware training (QAT) was performed using pseudo-labels generated via Homographic Adaptation, a technique originally proposed in the SuperPoint training procedure.
At the core of this method is a process that applies randomly sampled homographies to warped versions of the input image and aggregates the corresponding predictions.
To generate training annotations, a set of pseudo-ground-truth interest point locations was created for each image from the target domain, using the MS-COCO dataset \cite{ms-coco} as input. 
These annotations were then used to train the model using standard supervised learning techniques.
For models quantised to 8-bit and 4-bit precision, weights in floating-point format were used as input. In contrast, for 3-bit models and the 4-2-4 mixed precision variant, weights previously trained in 4-bit precision were used. This approach enabled the use of progressive quantisation, which contributes to increasing the stability of the learning process and improving the final accuracy of the model.

\subsection{Vitis AI}
\label{subsec:vitis_ai}

To utilise the DPU as a neural network accelerator, AMD Xilinx's Vitis AI tool was employed. It enables the necessary model optimisation and conversion steps required to run the model in the programmable logic resources.
The model preparation process consists of two main stages: quantisation and compilation.
The Vitis AI Quantizer tool was used to perform the quantisation process, converting network weights and activations from floating-point to INT8 fixed-point format. Alternatively, quantised weights can be obtained directly using the Brevitas library.
Vitis AI Compiler was then used to prepare the model for execution on the target hardware.
This tool analyses the model structure, identifies DPU hardware-supported operations, and performs layer merges, transformations, and simplifications to improve execution efficiency.
The model is divided by the compiler into subgraphs, each of which is assigned to a specific hardware resource.
DPU-supported operations are allocated to the PL part, while the rest are allocated to the PS, which is usually an embedded ARM processor.
This separation allows for efficient use of the available resources of a heterogeneous computing platform.

The \texttt{pynq-dpu} library, provided by AMD Xilinx, is used to run and transfer data between hardware resources. The DPU configuration bitstream DPUCZDX8G ISA0 B4096 MAX BG and the file containing the compiled neural network model are loaded with the help of the provided tool.
In a subsequent step, the data conversion between floating-point and fixed-point representations is performed, and the corresponding input and output buffers, corresponding to the tensor sizes, are created. Then asynchronous inference on the DPU is initiated.
Data sent to and from the DPU is transferred via buffers in shared memory, while the communication between the PS and the PL is carried out using the AXI interface. Finally, the resulting data is converted back to floating-point format.

\subsection{FINN}

FINN \cite{finn} is an AMD tool that supports the exploration of hardware implementations of deep neural networks.
Its core components are a compiler and a library of kernels that implement basic operations in reprogrammable logic.
The compilation process uses the ONNX graph representation, extended by the QONNX library \cite{qonnx_paper, qonnx_repo}, which introduces, among other features, a custom representation of quantisation operations for arbitrary bit widths. 
The compiler includes a set of transformations that, among other things, shift floating-point affine operations -- originating from quantisation or batch normalisation -- into forms more suitable for hardware implementation.

The activation function and the activation requantification operation are implemented using a thresholding operation in which the input $x$ is compared with a set of thresholds $T = t_0, t_1,.... t_n\}$.
The output is then the smallest index $i$ for which $t_i > x$.
In a sense, this is a computation of quantisation steps at the compilation stage instead of doing it at runtime using affine operations.
For each neural network topology, the order of reordering transformations should be chosen individually so that all floating-point affine operations $ax + b$ are just before the thresholding operation.
Then it is possible to avoid directly calculating these operations by substituting $t_i \gets (t_i - b)/a$.

After the compilation process, a suitable hardware kernel is selected for each node in the computational graph to create a pipelined processing architecture in reprogrammable logic.
The model parameters are stored in the RAM of the PS and are sent to the accelerator in PL via DMA.

For more details, we refer the reader to the publications of FINN authors \cite{finn, finnr}.

\section{Evaluation}
\label{sec:evaluation}


This section presents the evaluation of a quantised SuperPoint network on FPGA platforms using two approaches: Vitis AI and FINN. The Vitis AI implementation targets the DPU with 8-bit quantisation for high efficiency, while the FINN approach, using Brevitas, allows for custom low-bit quantisation and hardware-aware optimisation.


\subsection{Experiment setup}
The experiments were conducted using image sequences stored on an SD card, within a Jupyter Notebook environment. On the Kria KV260 and ZCU102 platforms, the PYNQ environment with a Python application was used. Both boards were connected to a PC via Ethernet. The DPU accelerator with Vitis AI was tested only on the Kria platform, as the DPU supports only 8-bit quantised models and its use is therefore limited. The FINN-based implementation, supporting various precision levels, was evaluated on both boards.

\subsection{Performance Analysis of the SuperPoint Detector}

A detailed analysis of network variants trained with low numerical precision was carried out using the Brevitas library. This analysis aimed to evaluate the impact of the applied quantisation on the quality of the SuperPoint network detector and descriptor performance.
Three metrics were used to evaluate the system: repeatability, localisation error, and homography estimation.
Detector Repeatability is used to evaluate the precision of feature point detection. Localisation Error measures the average location error of detected points relative to their actual locations. 
Homography Estimation evaluates the accuracy of descriptor matching by computing the error in estimating a homography between image pairs.
The experiments were carried out using the HPatches dataset \cite{hpatches_2017_cvpr}, a widely used benchmark to evaluate the performance of feature detectors and descriptors.
The obtained results are summarised in Table \ref{tab:detector_metrics} and compared with the results of other existing works on hardware implementation of SuperPoint networks on FPGA platforms. 
It should be noted that most of them modify the network architecture, simplifying it to accommodate hardware limitations.
In contrast to these approaches, the method proposed in this paper preserves the original network architecture. It focuses on using quantisation as a mechanism to reduce the number of parameters and improve the efficiency of the hardware implementation. This makes it possible to maintain high quality feature point detection while significantly reducing the demand for computational and memory resources, making the model more suitable for implementation on FPGA-type platforms. 
Across all tested quantisation levels, Repeatability and Localization Error remained largely unaffected.
However, larger differences are observed for Homography Estimation, indicating that this metric is more sensitive to changes in computational precision.
Despite the use of quantisation, the Repeatability metric achieves higher values than in the other works analysed. The Localisation Error remains at a comparable level, and the Homography Estimation accuracy exceeds the results of the two existing solutions \cite{xu2020, Yin2024} and is only slightly worse than the best of the compared models  \cite{liu2022}.

\begin{table}[t]
\centering
\caption{Comparison of SuperPoint (INT8) network execution times on the AMD Xilinx FPGA Kria KV260 platform and the average number of point matches on a set of TUM and KITTI depending on usage, depending on the location of the Softmax operation (DPU or PS -- Processing System).}
\label{tab:superpoint_dpu}
\resizebox{\columnwidth}{!}{%
    \begin{tabular}{cc|cc|cc|}
    \cline{3-6}
    \multicolumn{1}{l}{}                    & \multicolumn{1}{l|}{} & \multicolumn{2}{c|}{\textbf{PS Softmax}}    & \multicolumn{2}{c|}{\textbf{DPU Softmax}}    \\ \hline
    \multicolumn{1}{|c|}{\textbf{Image size}} &
      \textbf{Dataset} &
      \multicolumn{1}{c|}{\textbf{\begin{tabular}[c]{@{}c@{}}Frame \\ Rate [FPS] \end{tabular}}} &
      \textbf{\begin{tabular}[c]{@{}c@{}}Avg. \# of \\ matches\end{tabular}} &
      \multicolumn{1}{c|}{\textbf{\begin{tabular}[c]{@{}c@{}}Frame \\ Rate [FPS]\end{tabular}}} &
      \textbf{\begin{tabular}[c]{@{}c@{}}Avg. \# of \\ matches\end{tabular}} \\ \hline
    \multicolumn{1}{|c|}{$80 \times 60$}    & TUM                   & \multicolumn{1}{c|}{353.4}   & 13.25  & \multicolumn{1}{c|}{353.4}   & 8.17   \\ 
    \multicolumn{1}{|c|}{$160 \times 120$}  & TUM                   & \multicolumn{1}{c|}{119.2}   & 64.44  & \multicolumn{1}{c|}{148.8}   & 28.89  \\ 
    \multicolumn{1}{|c|}{$320 \times 240$}  & TUM                   & \multicolumn{1}{c|}{33.8}  & 221.27 & \multicolumn{1}{c|}{41.0}  & 88.41  \\ 
    \multicolumn{1}{|c|}{$640 \times 480$}  & TUM                   & \multicolumn{1}{c|}{8.7} & 551.30 & \multicolumn{1}{c|}{10.6}  & 194.55 \\ 
    \multicolumn{1}{|c|}{$416 \times 128$}  & KITTI                 & \multicolumn{1}{c|}{48.4}  & 123.41 & \multicolumn{1}{c|}{58.6}  & 82.37  \\ 
    \multicolumn{1}{|c|}{$1241 \times 376$} & KITTI                 & \multicolumn{1}{c|}{5.7} & 460.61 & \multicolumn{1}{c|}{7.0} & 199.00 \\ \hline
    \end{tabular}%
}
\end{table}

\begin{table}[t]
\centering
\caption{Comparison of network execution times and number of matches for the KITTI dataset and image size $416 \times 128$ px, depending on the location of the Softmax operation (DPU or PS -- Processing System) on the AMD Xilinx FPGA Kria KV260 platform.}
\label{tab:cpu_dpu_softmax}
\begin{tabular}{|c|c|c|c|c|}
\hline
\textbf{Configuration} &
  \textbf{\begin{tabular}[c]{@{}c@{}}Inference \\ time [ms]\end{tabular}} &
  \textbf{\begin{tabular}[c]{@{}c@{}}Softmax \\ time [ms]\end{tabular}} &
  \textbf{\begin{tabular}[c]{@{}c@{}}End-to-end \\ latency [ms]\end{tabular}} &
  \textbf{\begin{tabular}[c]{@{}c@{}}Avg. \# \\ of matches\end{tabular}} \\ \hline
PS Softmax & 16.86 & 3.80 & 20.66 & 123.41 \\ 
DPU Softmax & 17.07 & -    & 17.07 & 82.37  \\ \hline
\end{tabular}
\end{table}

\begin{table*}[t]
\centering
\caption{Performance comparison of SuperPoint accelerators.}
\label{tab:utilisation}
\begin{tabular}{l|c|c|c|c|c|c|}
\cline{2-7} & FCCM’20 \cite{xu2020} & TCASI’22 \cite{liu2022} & APCCAS’24 \cite{Yin2024} & \multicolumn{3}{c|}{Our work}  \\ \hline
\multicolumn{1}{|l|}{\textbf{Framework}}                    & Vitis AI (DPU)         & Vitis AI (DPU)         & Vivado HDL (PL)         & Vitis AI (DPU) & FINN (HLS)          & FINN (HLS)       \\
\multicolumn{1}{|l|}{\textbf{FPGA}}                         & ZCU102                & ZCU104                  & ZCU102                 & Kria KV260 & Kria KV260          & ZCU102         \\  \hline \hline
\multicolumn{1}{|l|}{\textbf{LUT}}                          & 80599                 & 100913                  & 22876                  &  52161* & 95046                & 157359         \\
\multicolumn{1}{|l|}{\textbf{FF}}                           & 175434                & 195029                  & 23623                 &   98249* & 142801         & 190952         \\
\multicolumn{1}{|l|}{\textbf{DSP}}                          & 1307                  & 1318                    & 145                    &  710* & 8                    & 8              \\
\multicolumn{1}{|l|}{\textbf{BRAM}}                         & 499.5                 & 205                     & 122                   &  255* & 76.5                 & 38.5           \\
\multicolumn{1}{|l|}{\textbf{Resolution}}                   & $640 \times 480$      & $640 \times 480$        & $640 \times 480$      & $640 \times 480$  & $640 \times 480$  & $640 \times 480$ \\
\multicolumn{1}{|l|}{\textbf{Frequency [MHz]}}              & 200                   & 150                     & 200                   &  200 & 300 & 300            \\
\multicolumn{1}{|l|}{\textbf{Frame Rate [FPS]}}             & 20                    & 42                      & 20                    &  10.6 & 27                   & 54             \\
\multicolumn{1}{|l|}{\textbf{Latency [ms]***}}             & -                    & -                      & -                    &  94 & 66                   & 33             \\
\multicolumn{1}{|l|}{\textbf{Energy Efficiency [mJ/frame]}} & -                     & -                       & 58.2                  & 188.7 -- 377.3  & 122.5                & 85.3           \\
\multicolumn{1}{|l|}{\textbf{Power [W]}} & - & - & 1.16 & \multicolumn{1}{c|}{2 -- 4*} & \multicolumn{1}{c|}{\begin{tabular}[c]{@{}c@{}}Total: 3.307** \\ PL: 0.890**\\ PS: 2.417**\end{tabular}} & \begin{tabular}[c]{@{}c@{}}Total: 4.607**\\ PL: 1.772**\\ PS: 2.835**\end{tabular} \\ \hline

\end{tabular}
\begin{flushleft}
* Parameters were given on the basis of documentation \cite{amd_dpu_resources}. \\
** Estimated with Vivado Power Analysis tool. \\
*** The term 'latency' refers to the time it takes to fully process one frame.
\end{flushleft}
\end{table*}

\subsection{Vitis AI implementation}

At the hardware implementation stage, the Vitis AI environment and DPU accelerator were first tested to evaluate several factors. These included the speed of operation, the quality of the generated descriptors through the efficiency of feature point matching, and the impact of where the Softmax function is implemented.
Two scenarios were considered for distributing the computational workload across the heterogeneous system's resources.
In both configurations, the main computational component -- involving successive convolutional layers of the neural network in INT8 format -- was implemented on the DPU, which is part of programmable logic.
The key difference between the scenarios concerned the way the Softmax function was implemented. 
Please refer to Table \ref{tab:superpoint_dpu} for a comparison of the processing times of the SuperPoint network running on the AMD Xilinx Kria KV260 platform and the average number of correct feature point matches for two datasets: TUM RGB-D and KITTI.
The summary considers various configurations of the implementation of the Softmax function, including those executed on the processor (PS) and those executed directly on the hardware accelerator (PL).
The impact of the limited precision of the calculations on the quality of results is also taken into account.
The presented data allows an assessment of the trade-off between computational efficiency and accuracy of detected and matched feature points.

In the first scenario, this operation was executed in software on the processor (PS). In the second scenario, it was quantised and integrated with the other operations within the PL thanks to hardware support for the Softmax function on the DPU.
This approach enabled a thorough analysis of the impact of moving the Softmax function to a hardware accelerator on the final performance and accuracy of the system.
Table \ref{tab:cpu_dpu_softmax} presents a comparison of the processing times of the neural network in the DPU and the number of correct feature point matches obtained using the SuperGlue network.
The evaluation demonstrated that the quantised version of the network generates 10 to 15\% fewer feature points compared to the full-precision implementation.
Although the DPU supports Softmax operations in hardware, these are performed in fixed-point precision (INT8), which may lead to a degradation in output quality, particularly in terms of the accuracy of point localisation and the number of correct matches. 
Consequently, the number of correct correspondences between images may be lower compared to the variant in which the Softmax function is calculated in the Processing System (PS) using floating-point precision.
On the other hand, offloading the Softmax operation to the DPU resulted in a performance improvement of approximately 18\%, enabling the network to achieve a throughput of 58.6 frames per second (FPS).

It should be noted that it is not possible to precisely determine the utilisation of programmable logic (PL) resources by a given model instance, due to the internal allocation mechanisms of the DPU. 
However, based on the selected configuration, in this case DPUCZDX8G ISA0 B4096 MAX BG, it is possible to estimate the approximate utilisation of PL resources, as summarised in Table \ref{tab:utilisation}.


\subsection{FINN implementation}

This section presents an analysis of the hardware resource utilisation, as well as the time and power efficiency of the proposed SuperPoint network implementations using FINN.
The objective of the analysis was to assess the efficiency of implementing the quantised model in FPGA resources.
Table \ref{tab:utilisation} summarises the utilisation of programmable logic resources, the processing performance achieved, and the estimated power consumption for two variants of the SuperPoint network implementations developed using the FINN framework.
These results are compared with selected state-of-the-art solutions reported in the literature.
Two levels of parallelism were considered: the first is designed for the Kria KV260 platform, and the second is designed for the larger ZCU102 development board. 
The former achieves a throughput of 27 frames per second (FPS), while the latter reaches 54 FPS.
Block RAM (BRAM) consumption is primarily due to the implementation of DMA buffers and thresholding modules, which store activation requantisation thresholds.
The power consumption values were obtained from Vivado Power Analysis tool and include both programmable logic (PL) and processing system (PS) components.
Although all the processing is done in the PL, the PS accounts for most of the power consumption (73\% for the Kria and 62\% for the ZCU102).
However, its role of PS crucial, as it is responsible for sending the network weights to the accelerator in the PL.
For the Kria platform, due to the limited availability of LUT resources, sliding window units responsible for generating convolutional contexts were implemented using BRAM.
It is important to note that the data presented in Table \ref{tab:utilisation} correspond to the 3-bit implementation, which was selected for detailed analysis.
The resource utilisation between the 4-2-4 and 3-bit variants was found to differ only slightly. 
Moreover, the performance quality of the detector in the 3-bit version remains competitive with other solutions described in the literature. 
These solutions achieve low resource requirements by simplifying the network architecture. 
Our implementation preserves the full structure of the SuperPoint model, which translates into better detection performance.

\begin{table}[!ht]
\centering
\caption{Comparison of trajectory estimation results on the TUM dataset for different levels of precision quantisation of the SuperPoint model using RPE and APE metrics.}
\label{tab:rpe_ape_metrics}
\begin{tabular}{|l|c|c|c|c|}
\hline
 &
  \multicolumn{1}{l|}{\textbf{R$_{RPE} [^{\circ}]$}} &
  \multicolumn{1}{l|}{\textbf{t$_{RPE} [m/s]$}} &
  \multicolumn{1}{l|}{\textbf{R$_{APE}[^{\circ}]$}} &
  \multicolumn{1}{l|}{\textbf{t$_{APE} [m]$}} \\ \hline
\multicolumn{1}{|l|}{\textbf{FCCM’20 \cite{xu2020}}}                                      & -       & 0.0283 & -       & 0.3976 \\
\textbf{\begin{tabular}[c]{@{}l@{}}Baseline\\ (FP32 \cite{detone2018superpoint}) \end{tabular}}            & 1.5765  & 0.0260 & 7.6082  & 0.3106 \\
\textbf{INT8}                                                                & 2.1714  & 0.0344 & 9.3949  & 0.3843 \\
\textbf{INT4}                                                                & 3.5987  & 0.1158 & 10.4052 & 0.3630 \\
\textbf{INT3}                                                                & 5.5383  & 0.1331 & 12.2407 & 0.3711 \\
\textbf{\begin{tabular}[c]{@{}l@{}}Mixed precision\\ (INT 4-2-4)\end{tabular}} & 10.3530 & 0.1603 & 19.9678 & 0.5316 \\ \hline
\end{tabular}
\end{table}

\subsection{Evaluation of Visual Odometry Performance}
\begin{figure}[!ht]
    \centering
    \includegraphics[width=1\linewidth]{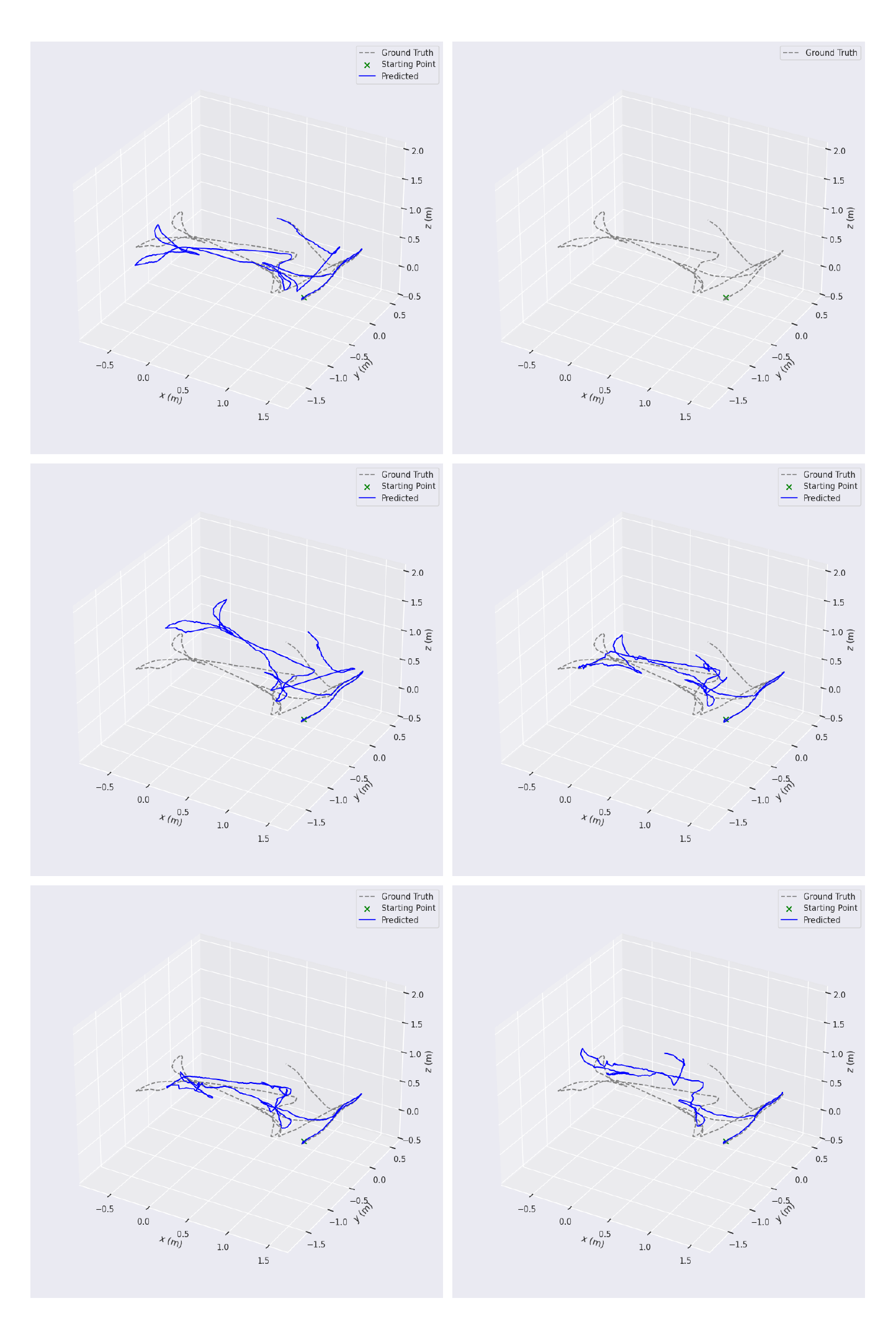}
    \caption{Comparison of estimated camera trajectories obtained using different quantisation levels of the SuperPoint network on the TUM dataset, with reference to the ground truth trajectory. First row: trajectories for the model at floating point precision and the reference trajectory (ground truth). Second line: trajectories for models quantised to 8 and 4 bits. Third row: trajectories for models quantised up to 3 bits and for mixed precision (4-2-4 bit).}
    \label{fig:trajectory_comparison}
\end{figure}

\begin{figure*}[!ht]
    \centering
    \includegraphics[width=1\textwidth]{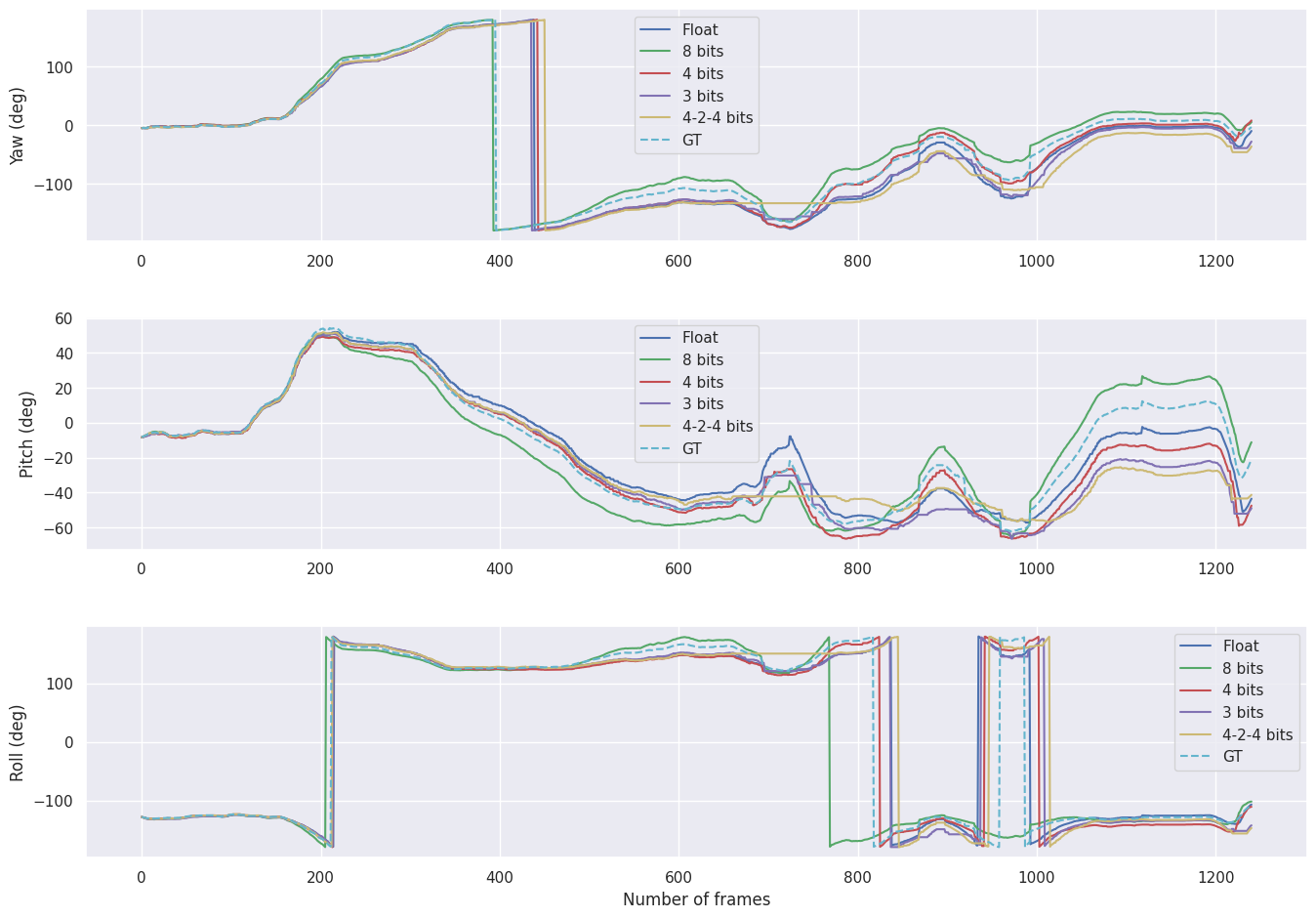}
    \caption{Comparison of estimated orientation angles (roll, pitch, yaw) with reference values (ground truth) for the trajectory recorded on the TUM dataset.}
    \label{fig:angles_comparison}
\end{figure*}

This section presents the results of evaluating the quality of the odometry system performance depending on the degree of SuperPoint network quantisation used. The impact of varying precision representations (8, 4, 3 bits and a mixed 4-2-4 configuration) on the accuracy of trajectory estimation was thoroughly analysed. Two metrics were selected for objective evaluation: 

\begin{itemize}
    \item APE (Absolute Pose Error) is a measurement of the global accuracy of the estimated trajectory relative to the reference data (ground truth). For each frame, the difference in absolute position and orientation relative to the reference trajectory is analysed.

    \item RPE (Relative Pose Error) assesses the local accuracy of the estimated trajectory by analysing the relative error between successive positions. This metric better reflects the quality of odometry, as it is not prone to error accumulation in long trajectories.
\end{itemize}
Both metrics report the Root Mean Square Error (RMSE) value, calculated independently for the translational, rotational, and full transformation components, SE (3).
As shown in Table \ref{tab:rpe_ape_metrics}, a comparison of results for different precision representations (8-bit, 4-bit, 3-bit, and mixed precision 4-2-4) on the TUM RGB-D dataset ($640 \times 480$ resolution) is presented. 
The evaluation was conducted using standard visual odometry metrics: Relative Pose Error (RPE) -- separately for RPE rotation (R$_{RPE} [^{\circ}]$) and translation (t$_{RPE} [m/s]$), and Absolute Pose Error (APE) -- also for APE rotation (R$_{APE}[^{\circ}]$) and APE translation (t$_{APE} [m]$).
It can be observed that decreasing the numerical precision leads to increased translation and rotation errors for both RPE and APE.
The largest increase in errors is seen for the mixed precision variant (INT 4-2-4), indicating the significant impact of quantisation on the accuracy of position and orientation estimation.
For comparison, the results of the full-precision baseline model (FP32) \cite{detone2018superpoint} and a reference implementation from the literature (FCCM’20 \cite{xu2020}) are also included.
For the FP32 baseline, the APE rotation error is 7.61$^\circ$, and the APE translation error is 0.31~m.
Quantisation to INT8 results in a moderate increase: R$_{APE}$ increases to 9.39$^\circ$ (approximately +23\%), while t$_{APE}$ rises to 0.38 m (approximately 24\% higher than FP32).
In the case of INT4, R$_{APE}$ reaches 10.41$^\circ$ (about 37\% higher than FP32), and t$_{APE}$ is 0.36 m, which is only slightly higher than the baseline.
Moving to INT3, the APE rotation error jumps to 12.24$^\circ$ (an increase of over 60\% compared to FP32), and the APE translation error is 0.37 m (still within a 20\% margin of FP32).
The most significant degradation is observed for the mixed precision (INT 4-2-4) variant, where the R$_{APE}$ rises to 19.97$^\circ$ -- over 2.6 times the error of the baseline model. t$_{APE}$ also increases markedly to 0.53 m, which is nearly 70\% higher than the FP32 baseline.
As illustrated in Figures \ref{fig:trajectory_comparison} and \ref{fig:angles_comparison}, the estimated orientation angles (roll, pitch, and yaw) are compared with the ground truth data.
In addition, the full motion trajectory for the TUM dataset is illustrated, enabling a qualitative assessment of the estimated path's accuracy relative to the actual one.
It is important to note that the presented results were obtained without any alignment or optimisation procedures applied to the estimated trajectories.
As shown in Figure \ref{fig:angles_comparison}, the deviation of the 8-bit quantised representation from the ground truth varies depending on the angle. However, from the perspective of the entire trajectory, the 8-bit quantised model provides the most accurate overall approximation. 
While in certain local segments the 4-bit or even 3-bit variants better capture specific angle transitions, these lower-precision models generally result in larger cumulative errors.
The mixed precision (INT 4-2-4) variant yields the poorest performance, as it fails to respond adequately to rapid changes in orientation, resulting in greater discrepancies from the ground truth, particularly during dynamic motion. 

Although the estimated trajectories broadly follow the shape of the reference path, discrepancies accumulate over time, leading to visible deviations from the ground truth.
Nevertheless, these errors could be mitigated through post-processing, such as trajectory alignment or global optimisation techniques.

\section{Conclusion}
\label{sec:discussion}


In this work, we presented an embedded implementation of a quantised SuperPoint neural network designed for unsupervised feature point detection and description. The solution was deployed on an AMD/Xilinx Zynq UltraScale+ FPGA System-on-Chip platform, where we evaluated the performance of Deep Learning Processing Units (DPUs). Using the Brevitas library and the FINN framework, we applied quantisation-aware training and hardware-specific optimisations. This enabled real-time processing of $640 \times 480$ images at up to 54 frames per second. We also conducted experiments on the TUM dataset to evaluate the impact of different quantisation strategies on both model accuracy and performance in a visual odometry context.
The streaming architecture of the SuperPoint accelerator, developed using the FINN framework, enabled significantly higher throughput compared to related FPGA-based implementations reported in the literature. Furthermore, the flexible allocation of hardware resources (LUTs, BRAMs, and DSPs) per accelerator module allows for efficient adaptation to the constraints of different target platforms, making the approach scalable and widely applicable in embedded computer vision systems.
Among the tested configurations, the most promising hardware implementation was achieved using 3-bit quantisation. 
Although this approach enables implementation on FPGA devices, achieving high accuracy and precise feature point detection remains challenging at such low precision. Nevertheless, the possibility of deploying 3-bit hardware quantisation demonstrates the potential for further optimisation of neural networks in embedded, resource-constrained environments.

In practical terms, a slight reduction in detection accuracy may be acceptable in scenarios where implementation on hardware with limited computational and energy resources is required. In such embedded applications, energy efficiency and minimal resource usage are often of primary importance, and moderate compromises in accuracy can be tolerated from the perspective of the overall system.

\begin{figure}[!ht]
    \centering
    \includegraphics[width=1\columnwidth]{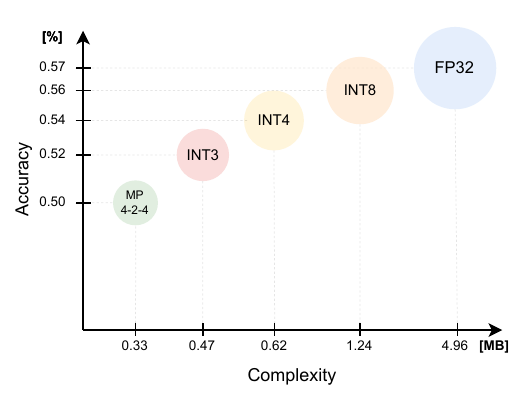}
    \caption{Trade-off between feature point detection accuracy and bitwidth (computational complexity) for different SuperPoint model variants.}
    \label{fig:pareto}
\end{figure}

Figure \ref{fig:pareto} illustrates the impact of different SuperPoint model variants (original and quantised to 8, 4, 3, and 4-2-4 bits) on the accuracy of the detection of feature points as a function of the number of network parameters. 
Detection accuracy values are taken from Table \ref{tab:detector_metrics}. 
Each point represents a different model variant, illustrating the trade-off between model complexity and detection performance.

While the practical benefits of preserving the full network architecture may not always be obvious, quantisation offers several key advantages.
Firstly, it does not require significant modifications to the original model structure, making the process straightforward and less error-prone.
Furthermore, quantisation of an existing model is generally more efficient, both in terms of time and computational resources, than designing, implementing, and training a new architecture from scratch.
Maintaining the original architecture also ensures compatibility with established training procedures, improves reproducibility, and facilitates direct comparisons with previously reported results.
This approach allows for a fair evaluation of quantisation effects and supports rigorous benchmarking against the baseline model.

Future research will focus on extending the current architecture to support full SLAM pipelines by integrating additional modules such as feature point matching and pose estimation directly on the FPGA. We also intend to explore mixed precision quantisation and sparsity techniques to further improve efficiency while maintaining model accuracy.

\bibliographystyle{IEEEtran} 
\bibliography{dsd_2025_superpoint} 

\end{document}